\documentclass[conference]{IEEEtran}
\usepackage[absolute,showboxes,overlay]{textpos}
\usepackage{hyperref}

\TPMargin{5pt}

\newcommand{\copyrightstatement}{
    \begin{textblock}{0.84}(0.08,0.93)    
         \noindent
         \footnotesize
         \copyright 2021 IEEE. Personal use of this material is permitted. Permission from IEEE must be obtained for all other uses, in any current or future media, including reprinting/republishing this material for advertising or promotional purposes, creating new collective works, for resale or redistribution to servers or lists, or reuse of any copyrighted component of this work in other works. {Cite from IEEE: }\href{<https://ieeexplore.ieee.org/document/9653015>}{DOI No. 10.1109/MILCOM52596.2021.9653015}
    \end{textblock}
}

\IEEEoverridecommandlockouts
\usepackage{cite}
\usepackage{amsmath,amssymb,amsfonts}
\usepackage{graphicx}
\usepackage{textcomp}
\usepackage{xcolor}

\usepackage{subfigure}
\usepackage{fancyhdr}
\usepackage{algorithm}
\usepackage{algpseudocode}
\usepackage{booktabs} 
\usepackage{multirow}

\include{pythonlisting}

\def\BibTeX{{\rm B\kern-.05em{\sc i\kern-.025em b}\kern-.08em
    T\kern-.1667em\lower.7ex\hbox{E}\kern-.125emX}}
\begin{document}
\copyrightstatement

\title{Deep GEM-Based Network for Weakly Supervised UWB Ranging Error Mitigation}

\author{
\IEEEauthorblockN{
Yuxiao~Li\IEEEauthorrefmark{1},
Santiago~Mazuelas\IEEEauthorrefmark{2}, and
Yuan~Shen\IEEEauthorrefmark{1}}
\IEEEauthorblockA{\IEEEauthorrefmark{1}
Department of Electronic Engineering,
Tsinghua University,
Beijing, China \\
}
\IEEEauthorblockA{\IEEEauthorrefmark{2}
BCAM-Basque Center for Applied Mathematics, and IKERBASQUE-Basque Foundation for Science, Bilbao, Spain \\
Email: li-yx18@mails.tsinghua.edu.cn,
smazuelas@bcamath.org,
shenyuan\_ee@tsinghua.edu.cn
}}

\maketitle

\begin{abstract}

Ultra-wideband (UWB)-based techniques, while becoming  mainstream approaches for high-accurate positioning, tend to be challenged by ranging bias in harsh environments. The emerging learning-based methods for error mitigation have shown great performance improvement via exploiting high semantic features from raw data. However, these methods rely heavily on fully labeled data, leading to a high cost for data acquisition. 
We present a learning framework based on weak supervision for UWB ranging error mitigation.
Specifically, we propose a deep learning method based on the generalized expectation-maximization (GEM) algorithm for robust UWB ranging error mitigation under weak supervision. Such method integrate probabilistic modeling into the deep learning scheme, and adopt weakly supervised labels as prior information. Extensive experiments in various supervision scenarios illustrate the superiority of the proposed method.

\end{abstract}

\begin{IEEEkeywords}
UWB radio, ranging error mitigation, weakly supervised Learning, generalized expectation-maximization algorithm, deep learning
\end{IEEEkeywords}

\section{Introduction}
\label{sec:intro}


Location-awareness has been playing an increasingly essential role in the new generation of wireless networks \cite{WinSheDai:J18,WinDaiShe:J18}, wherein centimeter-level precise positioning is required. Among the related approaches, Ultra-wideband (UWB)-based technique has continued too attract most of the research interest due to the wide bandwith of more than $500$ MHz in the $3.1-10.6$ GHz band \cite{WinSch:J02}.
However, its performance is often degraded in harsh environments due to multipath effects and non-line-of-sight (NLOS) conditions \cite{JohShuPet:J07}. 

Extensive error mitigation techniques have been proposed based on both statistical models and learning techniques \cite{DamDavWin:J08}. 
\begin{figure}[ht]
    \begin{center}
        \includegraphics[width=0.4\textwidth]{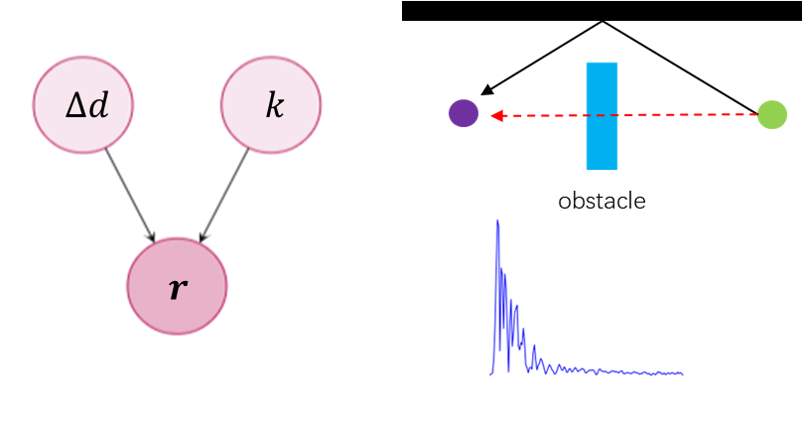}
        \caption{Illustrations of the proposed generative model for received signals. Instead of only considering mapping between signal and range measurement, we add in the environment label as the latent variable.}
        \label{fig:graphic}
    \end{center}
\end{figure}
Deep learning (DL) methods, 
emerging to be a popular trend recently, learn high semantic features from signals efficiently and in turn can achieve significant performance improvements  \cite{KleMih:J18,HaoOzgMik:J19,AngMazSalFanChi:J20}. 
These methods require large amount of labeled data, where both the received signals and their corresponding ranging error are known \cite{LiMazShe:C22,LiMazShe:C22_2}. As a result, the scarcity of labeled data has become a severe issue for developing efficient and robust solutions. 

In contrast to the labeled data, unlabeled data are much easier to obtain while also convey helpful modeling information \cite{Zhou:J18}. Several semi-supervised learning techniques have been developed in the world of wireless network applications, including Wi-Fi based localization \cite{YeHuaWan:J19}
and tracking mobile users \cite{JunJiaJie:J12}, while similar approaches have seldom claimed in UWB signal processing. These works develop efficient and relatively more robust learning solutions, which both extract high semantic features and require less labeling effort by exploiting both labeled and unlabeled data.

We consider a border scenario with respect to data supervision for UWB ranging error mitigation, known as weak supervision \cite{Zhou:J18,MazPer:C19}. Such scenario includes incomplete, inexact, and inaccurate labeling for data samples, which often occur at the same time in data acquisition.
We propose a deep neural network based on the generalized expectation-maximization (GEM) algorithm for UWB ranging error mitigation, which enables weakly supervised learning with coarsely labeled training data. 
The observed waveform together with the unobserved environment label are viewed as the complete data in the statistic model. The ranging error, accordingly, is modeled as the unknown parameter obtained by the maximum likelihood estimate (MLE) over the complete data. The weakly supervised labels (i.e., incomplete or coarse), in turns, are modeled to provide prior information. Two sub neural modules, referred to as E-Net and M-Net, are adopted to conduct the GEM algorithm in an end-to-end manner. During training, E-Net estimate the environment label while M-Net utilize raw received signal as well as the estimation from E-Net to accomplish the ranging error estimation.



The remaining sections are organized as follows. Section \ref{sec:model} introduces the problem statement and the proposed GEM framework. Section \ref{sec:network} introduces the implementation of the proposed method in deep learning.
Experimental results on two different datasets are illustrated in Section \ref{sec:exp}.
Finally, a conclusion and future focus can be found in Section \ref{sec:con}.

\section{Model Formulation}
\label{sec:model}

\subsection{Problem Statement}
\label{sec:problem}

In a harsh environment with obstacles and reflecting surfaces, the received signal at the agent can be written as follows,
\begin{equation}
\label{eq:state}
    \mathbf{r}(t) = \sum_{l} \alpha_l \mathbf{s}(t-\tau_l) + \mathbf{z}(t), ~t\in [0, T]
\end{equation}
\noindent where $\mathbf{s}(t)$ is a known wideband waveform, $\alpha_l$ and $\tau_l$ are the amplitude and delay, respectively, of the $l$th path, $\mathbf{z}(t)$ is the observation noise, and $[0, T]$ is the observation interval. We will denote $\mathbf{r}(t)$ as $\mathbf{r}$ for convenience in the rest of the paper. The relationship between the true distance $d$ and the delays of the propagation paths is:
\begin{equation}
    \tau_l =1/{c}(d + b_l)
\end{equation}
\noindent where $c$ is the propagation speed of the signal, and $b_l\geq 0$ is a range bias. Mostly, the range bias $b_l=0$ for LOS propagation, whereas $b_l>0$ for NLOS propagation. Suppose $d_{\text{M}}$ is the measured distance by the UWB device, the target of mitigation is to 
estimate the range bias in the specific path and remove from the measurement $d_{\text{M}}$.

Suppose the ranging error is denoted by $\Delta d$, where $\Delta d = d_{\text{M}}-d$. In the following we will show a GEM framework for efficient learning of the estimation of ranging error $\Delta d$ given the received signal $\mathbf{r}$.

\subsection{GEM Framework}

We take the actual ranging error $\Delta d$ as the unknown parameter to be estimated. From an aspect of MLE, the target is to estimate $\Delta d$ that maximizes $\log p(\Delta d|\mathbf{r})$. However, such distribution is hard to obtain due to the complicated propagation environment. Instead, we introduce the environment label $k$ for the latent variable, and estimate $p(k|\mathbf{r})$ together with $p(\Delta d|k, \mathbf{r})$ alternatively. The procedures are conducted by the GEM algorithm. Such environment label can be the LOS or NLOS conditions, different geometric rooms, or different blocking materials for the received signal. The MLE of $\Delta d$ is then conducted on complete data $(\mathbf{r}, k)$, i.e., $\log p(\Delta d|\mathbf{r}, k)$.


With the complete data being $(\mathbf{r}, k)$ and unknown parameter being $\Delta d$, the estimation of ranging error can be obtained from the MLE of the parameter by maximizing the conditional distribution of the observed data $\mathbf{r}$, written as:
\begin{equation}  \label{eq:obj}
    \begin{aligned}
        \log p(\mathbf{r}|\Delta d) \geq & \mathbb{E}_{q(k)}\big[\log\frac{p(\Delta d|k,\mathbf{r})}{p(\Delta d)}\big] - \operatorname{D}_{KL}\big(q(k)\big|\big|p(k|\mathbf{r})\big)  \\
        &+ \log p(\mathbf{r}) \\
        :=& \mathcal{F}(q,\Delta d;\mathbf{r})
    \end{aligned}
\end{equation}
\noindent where $\operatorname{D}_{KL}$ is the Kullback-Leibler (KL) divergence. The inequality in the second line is obtained from the Jensen's inequality, achieving equality iff $q(k)=p(k|\mathbf{r})$. 

The GEM algorithm seeks to find the estimation by iteratively applying the following two steps:

\begin{itemize}
    \item Expectation step (E-step)
    \begin{equation}
        q^{(n)} = \arg\max_q \mathcal{F}(q,{\Delta d}^{(n)};\mathbf{r})
    \end{equation}
    
    \item Maximization step (M-step)
    \begin{equation}
        {\Delta d}^{(n+1)} = \arg\max_{\Delta d} \mathcal{F}(q^{(n)},\Delta d;\mathbf{r})
    \end{equation}
\end{itemize}

To fulfill the formulation of the objective function in Eq.\eqref{eq:obj}, $p(\Delta d)$ and $p(k|\mathbf{r})$ can be given by prior knowledge, while expressions for $p(\Delta d|k,\mathbf{r})$ and $q(k)$ are required. Since these distributions are hard to be approximated by model knowledge, we adopt techniques from deep learning to accumulate knowledge from data. Specifically, we utilize neural networks as well as datasets labeled with actural ranging error as environment labels to learn their analytical forms. 

\begin{figure*}[htbp]
      \centerline{
      \includegraphics[width=0.8\textwidth]{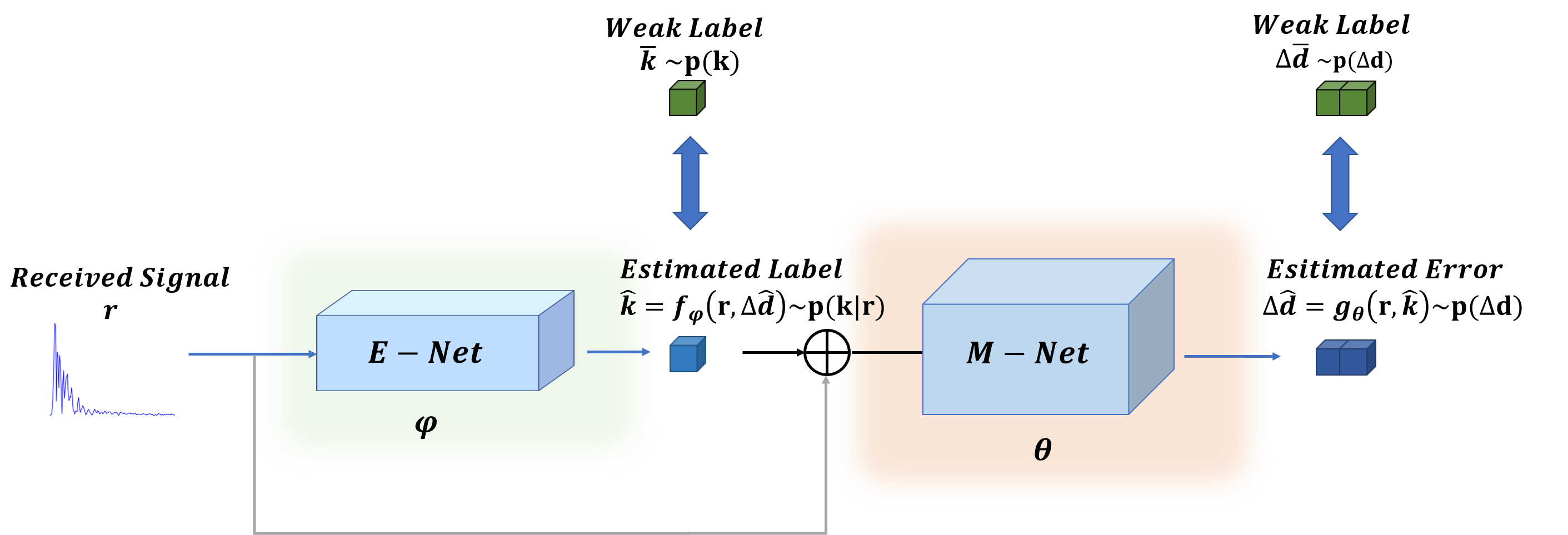}
      }
      \caption{Network structure of the proposed method, consisting of an E-Net to approximate the distribution of unobserved environment label and a M-Net to estimate the unknown ranging error. The network parameters, $\boldsymbol{\phi}$ and $\boldsymbol{\theta}$ respectively, are guided by the GEM algorithm on a weakly labeled dataset.}
      \label{fig:str}
  \end{figure*}

\section{Deep Learning Implementation}
\label{sec:network}

In this section, we construct a neural network to learn the optimization for the two steps. The network structure is illustrated in Fig.\ref{fig:str}.

\subsection{Weakly Labeled Dataset}

Suppose we are given a weakly labeled dataset $\mathcal{D}=\{\mathbf{r}^{(i)}, \bar{k}^{(i)}, \Delta \bar{d}^{(i)}\}_{i=1}^N$ with $N$ i.i.d. sample pairs, where $\bar{k}^{(i)}$ denotes the label for $i$th environment index, and $\Delta \bar{d}^{(i)}$ denotes the label for $i$th ranging error. Both labels are of weak supervision. In particular, these labels are coarsely gained and not always ground-truth. The specific values of these labels could be ground-truth with noise, randomly mismatched values for other samples, and at default. 
We take these weak labels to construct prior knowledge for the GEM framework.

Let $\hat{k}^{(i)}$ and $\Delta \hat{d}^{(i)}$ denote the according estimated environment label and ranging error.

\subsection{Neural Modules}

In E-step, the optimization of $q$ is conducted via learning $\hat{k}$ by the E-Net with parameter $\boldsymbol{\phi}$, i.e.,

\begin{equation}
    \hat{k}^{(i)} = f_{\boldsymbol{\phi}}(\mathbf{r}^{(i)}) \sim q(k)
\end{equation}
\noindent where $f_{\boldsymbol{\phi}}(\cdot):\mathbf{r}\rightarrow k$ denotes a vector-valued function parameterized by $\boldsymbol{\phi}$, mapping from the observed data $\mathbf{r}$ to the latent data $k$.

In M-step, the estimation of the ranging error $\Delta \hat{d}$ is obtained by the M-Net with parameter $\boldsymbol{\theta}$, i.e.,
\begin{equation}
    {\Delta \hat{d}}^{(i)}=g_{\boldsymbol{\theta}}(\mathbf{r}^{(i)}, \hat{k}^{(i)})
\end{equation}
\noindent where $g_{\boldsymbol{\theta}}(\cdot):\mathbf{r}\rightarrow \Delta d$ denotes a vector-valued function parameterized by $\boldsymbol{\theta}$, mapping from the observed data $\mathbf{r}$ to the unknown parameter $\Delta d$.

Merging into a whole end-to-end learning scheme, the objective function of the network with three neural modules can be expressed as:

\begin{equation}  \label{eq:net1}
    \boldsymbol{\phi},\boldsymbol{\theta} = \arg\max_{\boldsymbol{\phi},\boldsymbol{\theta}}\bar{\mathcal{F}}(\boldsymbol{\phi},\boldsymbol{\theta};\mathbf{r}).
\end{equation}

\subsection{Parametric Objective Function}


The optimization of the first expectation term $\mathbb{E}_{q(k)}\big[\log\frac{p(\Delta d|k,\mathbf{r})}{p(\Delta d)}\big]$ can be conducted via a MSE loss between ranging errors given
\begin{equation}  \label{eq:exp_par3}
    \begin{aligned}
        \mathcal{L}_{\text{exp}}(\boldsymbol{\varphi},\boldsymbol{\theta};\mathbf{r}^{(i)}, \Delta \bar{d}^{(i)}) &= \Vert \Delta \bar{d}^{(i)} - \Delta \hat{d}^{(i)} \Vert^2  \\
        &= \Vert \Delta \bar{d}^{(i)} - \Delta g_{\boldsymbol{\theta}}(k^{(i)},\mathbf{r}^{(i)}) \Vert^2
    \end{aligned}
\end{equation}

The optimization of the second KL term $-\operatorname{D}_{KL}\big(q(k)\big|\big|p(k|\mathbf{r})\big)$ can be conducted by the cross-entropy loss between label distributions given as
\begin{equation}  \label{eq:kl_para}
    \begin{aligned}
        \mathcal{L}_{kl}(\boldsymbol{\varphi};\mathbf{r}^{(i)}, \bar{k}^{(i)})  =& -\sum_{j=1}^K p(k_j^{(i)})\log q_{\boldsymbol{\varphi}}(k_j|\mathbf{r}^{(i)})  \\
        =& \operatorname{H}(p(k^{(i)})||q(k|\mathbf{r}^{(i)}))
    \end{aligned}
\end{equation}
\noindent where the variational distribution $q_{\boldsymbol{\varphi}}(k^{(i)})$ is learned by the network with parameter $\boldsymbol{\varphi}$, which can be simply done by empirically calculating the frequency of the output of the. The prior $p(k^{(i)}|\mathbf{r})$ is estimated empirically from the weak labels $\bar{k}^{(i)}$ from the dataset.

Therefore, we achieve the analytical version of the GEM objective in Eq.\ref{eq:obj} by combining Eqs.\eqref{eq:kl_para}-\eqref{eq:exp_par3}, which is differentiable w.r.t. parameters $\boldsymbol{\varphi}$ and $\boldsymbol{\theta}$ for the back-propagation (BP) algorithm for network learning.

\begin{equation}  \label{eq:obj}
    \begin{aligned}
    \bar{\mathcal{F}}(\boldsymbol{\varphi},\boldsymbol{\theta};\mathbf{r}^{(i)}, \bar{k}^{(i)}, \Delta \bar{d}^{((i))}) =& \mathcal{L}_{\text{exp}}(\boldsymbol{\varphi},\boldsymbol{\theta};\mathbf{r}^{(i)}, \Delta \bar{d}^{(i)})  \\
    &+ \mathcal{L}_{\text{kl}}(\boldsymbol{\varphi}; \mathbf{r}^{(i)}, \bar{k}^{(i)})
    \end{aligned}
\end{equation}

The optimization on dataset $\{\mathbf{r}^{(i)}, \bar{k}^{(i)}, {\Delta \bar{d}}^{(i)}\}_{i=1}^N$ can be conducted with the network structure in Fig.\ref{fig:str}, expressed as:

\begin{equation}  \label{eq:op}
    \boldsymbol{\varphi},\boldsymbol{\theta} = \arg\max_{\boldsymbol{\varphi},\boldsymbol{\theta}}\sum_{i=1}^N\bar{\mathcal{F}}(\boldsymbol{\varphi},\boldsymbol{\theta};\mathbf{r}^{(i)}, \bar{k}^{(i)}, \Delta \bar{d}^{(i)}).
\end{equation}

\section{Experiments}
\label{sec:exp}

\begin{table}[t]
\caption{Quantitative results of methods under different supervision rate $\eta_{\text{k}}$ as freezing $\eta_{\text{e}}=0.8$.}
\label{tab:exp_weak1}
\begin{center}
\begin{small}
\begin{sc}
\begin{tabular}{l|ccr}
\toprule
Methods & RMSE (m) & MAE (m) & Time (ms) \\
\midrule
Unmitigated & 0.428 & 0.291 & -  \\
SVM \cite{WymMarGifWin:J12} & 0.286& 0.175 & 4.915  \\

\midrule
\textit{GEM ($\eta_{\text{k}}=0.4$)}   & 0.135 & 0.074 & 1.643  \\
\textit{GEM ($\eta_{\text{k}}=0.6$)}   & 0.132 & 0.073& 2.368  \\
\textit{GEM ($\eta_{\text{k}}=0.8$)}   & 0.134 & 0.072 & 2.621  \\
\textit{GEM ($\eta_{\text{k}}=1.0$)}   & 0.123 & 0.072 & 0.983  \\
\bottomrule
\end{tabular}
\end{sc}
\end{small}
\end{center}
\end{table}

\begin{table}[t]
\caption{Quantitative results of methods under different supervision rate $\eta_{\text{e}}$ as freezing $\eta_{\text{k}}=0.8$.}
\label{tab:exp_weak2}
\begin{center}
\begin{small}
\begin{sc}
\begin{tabular}{l|ccr}
\toprule
Methods & RMSE (m) & MAE (m) & Time (ms) \\
\midrule
Unmitigated & 0.428 & 0.291 & -  \\
SVM \cite{WymMarGifWin:J12} & 0.286& 0.175 & 4.915  \\

\midrule
\textit{GEM ($\eta_{\text{e}}=0.4$)}   & 0.288 & 0.167 & 2.122  \\
\textit{GEM ($\eta_{\text{e}}=0.6$)}   & 0.220 & 0.122& 1.999  \\
\textit{GEM ($\eta_{\text{e}}=0.8$)}   & 0.134 & 0.072 & 2.621  \\
\textit{GEM ($\eta_{\text{e}}=1.0$)}   & 0.109 & 0.056 & 2.045  \\
\bottomrule
\end{tabular}
\end{sc}
\end{small}
\end{center}
\end{table}

\begin{figure*}[htbp]
    \begin{center}
        \subfigure[]{
        \begin{minipage}[t]{0.45\linewidth}
        \centerline{\includegraphics[width=0.9\textwidth]{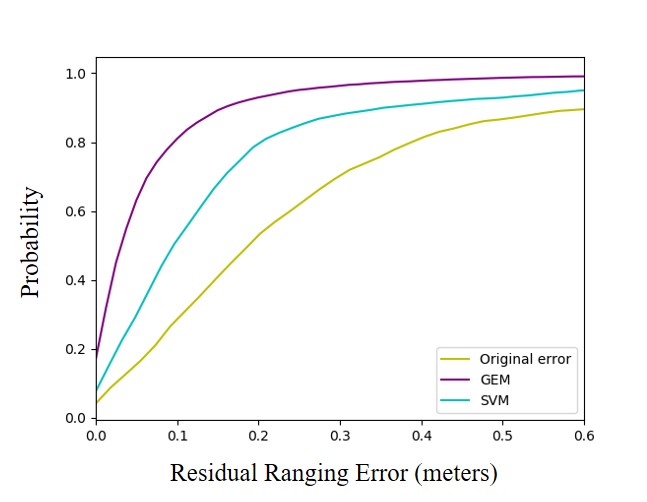}}
        \end{minipage}%
        }
        \subfigure[]{
        \begin{minipage}[t]{0.45\linewidth}
        \centerline{\includegraphics[width=0.9\textwidth]{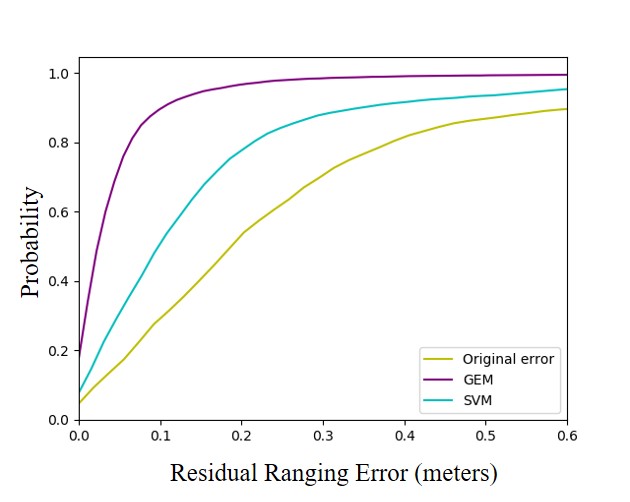}}
        \end{minipage}
        }
        \caption{The CDFs of the residual errors (remaining errors in range measurements after mitigation) after different mitigation methods under supervision rate (a) $\eta_{\text{k}}=0.4, \eta_{\text{e}}=0.8$, and (b) $\eta_{\text{k}}=0.8, \eta_{\text{e}}=0.8$. It can be seen that the proposed method outperforms SVM in both cases. The performance of the proposed method is rather robust to the supervision of environment label $k$, with a slight improvement with higher supervision rate $\eta_{\text{k}}$.}
        \label{fig:CDFs_e}
    \end{center}
\end{figure*}
\begin{figure*}[htbp]
    \begin{center}
        \subfigure[]{
        \begin{minipage}[t]{0.45\linewidth}
        \centerline{\includegraphics[width=0.9\textwidth]{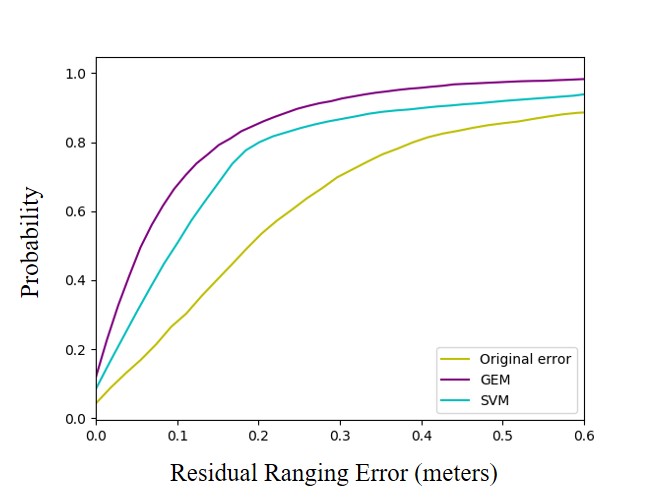}}
        \end{minipage}%
        }
        \subfigure[]{
        \begin{minipage}[t]{0.45\linewidth}
        \centerline{\includegraphics[width=0.9\textwidth]{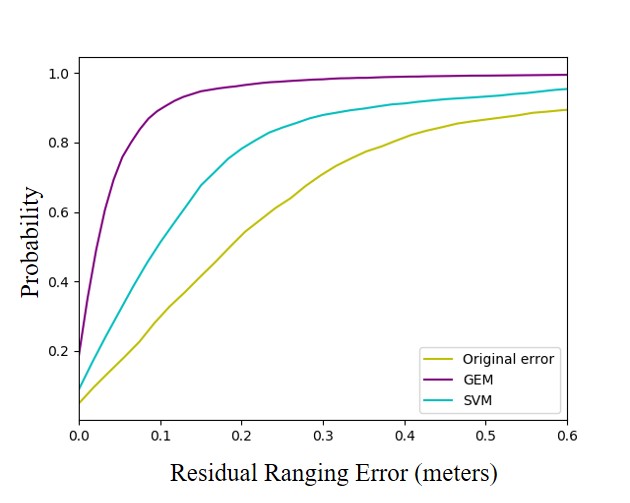}}
        \end{minipage}
        }
        \caption{The CDFs of the residual errors (remaining errors in range measurements after mitigation) after different mitigation methods under supervision rate (a) $\eta_{\text{k}}=0.8, \eta_{\text{e}}=0.4$, and (b) $\eta_{\text{k}}=0.8, \eta_{\text{e}}=0.8$. It can be seen that the proposed method outperforms SVM in both cases, while its performance is significantly improved with higher supervision rate $\eta_{\text{e}}$ for ranging error labels.}
        \label{fig:CDFs_m}
    \end{center}
\end{figure*}

The proposed method, referred to as GEM for convenience, is discussed with different weak supervision scenarios.
The values of the environment label are $k\in\{0, 1\}$ in our case, where $0$ refers to the LOS condition and $1$ refers to the NLOS condition.

\subsection{Database}

We compare the performance of our models with other methods on a public UWB database \cite{KleMih:J18}, consisting of the received waveforms, LOS or NLOS condition labels, and the actual ranging errors recorded in different indoor environments. The dataset is created using SNPN-UWB board with DecaWave DWM1000 UWB pulse radio module and generated in two different office environments. In the first environment, two adjacent office rooms with connecting hallway is considered, where $4800$ measurements in the first room and $5100$ measurements in the second. The second environment was a different office environment where multiple rooms, including $25100$ measurements in total. The waveform is represented as the absolute value of CIR, with the length of $152$. 
We assign $80\%$ of the data samples for training and the rest $20\%$ for testing, without overlapping between the two sets. 

\subsection{Data Processing and Baseline}

We test the algorithm under both weak and full supervision. For the full supervision case, the fully labeled dataset from the database is used, i.e. $\mathcal{D}_{\text{full}}=\{\mathbf{r}^{(i)}, \bar{k}^{(i)}, \Delta \bar{d}^{(i)}\}_{i=1}^N$ with $N$ i.i.d. sample pairs, where $\bar{k}^{(i)}$ denotes the actual label for $i$th environment index, and $\Delta \bar{d}^{(i)}$ denotes the actual label for $i$th ranging error.

For the weak supervision case, we synthesize a weakly labeled dataset $\mathcal{D}_{\text{weak}}$ from $\mathcal{D}_{\text{full}}$. Specifically, suppose the dataset consists of $N$ labeled samples $M_{\text{k}}$ samples with weak environment labels,  and $M_{\text{e}}$ samples with weak error labels. The weak label here refers to incomplete, inexact, or inaccurate cases. We define the supervision rate of environment label $k$ and error label $\Delta d$ as $\eta_{\text{k}}=\frac{N}{M_{\text{k}} + N}, \eta_{\text{e}}=\frac{N}{M_{\text{e}} + N}$

We randomly pollute the data labels with $\eta_{\text{k}}$ and $\eta_{\text{e}}$ by deleting, adding noises, and substituting values with other labels. 
The proposed method is evaluated under different supervision rates, i.e., $\eta_{\text{k}},\eta_{\text{e}}\in\{0.2, 0.4, 0.6, 0.8, 1.0\}$.

The classic Support Vector Machine (SVM) method is utilized as baseline method for ranging error mitigation, trained on the full supervised dataset with physical features extracted from the waveform, as suggested in \cite{MazConAllWin:J18,WymMarGifWin:J12}. It can be seen that the proposed method conducts efficient error mitigation under different supervision rates, and still outperforms SVM even with weak supervision.

\subsection{Results under Different $\eta_{\text{k}}$}

While supervision rate $\eta_{\text{e}}=0.8$ is frozen, the proposed method implemented with different supervision rate $\eta_{\text{k}}$ are compared. Quantitative results are shown in Table \ref{tab:exp_weak1}, in terms of root mean square error (RMSE), the mean absolute error (MAE), and inference time. It can be seen that methods under all supervision rates successfully mitigate the ranging error to some extend. Methods with the higher $\eta_{\text{k}}$ achieves better performance in error mitigation, while the performance rise w.r.t. $\eta_{\text{k}}$ is not tremendous. This implies that the proposed method can efficiently generate information from unlabeled data samples, especially the inherent information in environment label $k$. Thus, the proposed method can achieve a satisfactory performance with a more simple dataset weakly labeled in $k$ with a rate at around $0.4$.

\subsection{Results under Different $\eta_{\text{e}}$}

While supervision rate $\eta_{\text{k}}=0.8$ is frozen, the proposed method implemented with different supervision rate $\eta_{\text{e}}$ are compared. Quantitative results are shown in Table \ref{tab:exp_weak2}.
It can be seen that
methods with the higher $\eta_{\text{e}}$ achieves better performance in error mitigation, while the performance rise w.r.t. $\eta_{\text{e}}$ is more obvious compared to $k$. This implies that the proposed method can efficiently generate information from unlabeled data samples for ranging error information. However, the method is more sensitive to the supervision of ranging error $\Delta d$ than $k$. Thus, the proposed method can achieve a satisfactory performance with a more simple dataset weakly labeled in $\Delta d$ with a rate at around $0.6$.

It is worth noting that, almost all the results of the proposed method outperform SVM. This indicates the superiority of learning-based features to hand-crafted features for ranging error mitigation. In addition, the proposed method can exploit the weakly labeled dataset efficiently, while SVM requires fully labeled dataset.

\subsection{CDF Plots for Residual Ranging Error}

We additionally compare the ranging error mitigation performance in terms of the cumulative distribution function (CDF) for residual ranging errors (i.e., the remaining errors in range measurements after mitigation) under different supervision rates, illustrated in Fig.\ref{fig:CDFs_e}-\ref{fig:CDFs_m}.



By comparison between the two figures, the proposed approach achieves good results in both cases, while appears to be more sensitive to the supervision on $\eta_{\text{e}}$ than $\eta_{\text{k}}$. This phenomenon is consistent with the intuition that environment label $k$ takes place as a latent variable to give extra modeling information, while ranging error label $\Delta d$ serves as the ultimate estimation target.

\section{Conclusion}
\label{sec:con}

We proposed a weakly supervised learning approach based on GEM algorithm for UWB ranging error mitigation. The approach embedded the signal propagation model in a Bayesian framework, and enabled both efficient and robust estimation of the ranging error. 
Although 
proposed for UWB techniques, it provides a promising methodology for embedding Bayesian modeling in DL techniques, potential to benefit a wide range of learning problems involving a complicated process with latent variables. Future work would be focused on a more flexible framework on radio signal processing, integrating multiple related tasks in a unified Bayesian model.

\section*{Acknowledgment}
This research is partially supported by National Key R$\&$D Program of China 2020YFC1511803, the  Basque Government through the ELKARTEK programme, the Spanish Ministry of Science and Innovation through Ramon y Cajal Grant RYC-2016-19383 and Project PID2019-105058GA-I00, and Tsinghua University - OPPO Joint Institute for Mobile Sensing Technology.

\bibliographystyle{IEEEtran}
\bibliography{IEEEabrv,StringDefinitions,SGroupDefinition,refs}

\begin{thebibliography}{10}
\providecommand{\url}[1]{#1}
\csname url@samestyle\endcsname
\providecommand{\newblock}{\relax}
\providecommand{\bibinfo}[2]{#2}
\providecommand{\BIBentrySTDinterwordspacing}{\spaceskip=0pt\relax}
\providecommand{\BIBentryALTinterwordstretchfactor}{4}
\providecommand{\BIBentryALTinterwordspacing}{\spaceskip=\fontdimen2\font plus
\BIBentryALTinterwordstretchfactor\fontdimen3\font minus
  \fontdimen4\font\relax}
\providecommand{\BIBforeignlanguage}[2]{{%
\expandafter\ifx\csname l@#1\endcsname\relax
\typeout{** WARNING: IEEEtran.bst: No hyphenation pattern has been}%
\typeout{** loaded for the language `#1'. Using the pattern for}%
\typeout{** the default language instead.}%
\else
\language=\csname l@#1\endcsname
\fi
#2}}
\providecommand{\BIBdecl}{\relax}
\BIBdecl

\bibitem{WinSheDai:J18}
M.~Z. Win, Y.~Shen, and W.~Dai, ``\textnormal{A theoretical foundation of
  network localization and navigation},'' \emph{Proc. {IEEE}}, vol. 106, no.~7,
  pp. 1136--1165, Jul. 2018.

\bibitem{WinDaiShe:J18}
M.~Z. Win, W.~Dai, Y.~Shen, G.~Chrisikos, and H.~{Vincent Poor},
  ``\textnormal{Network operation strategies for efficient localization and
  navigation},'' \emph{Proc. {IEEE}}, vol. 106, no.~7, pp. 1224--1254, Jul.
  2018.

\bibitem{WinSch:J02}
M.~Z. Win and R.~Scholtz, ``Characterization of ultra-wide bandwidth wireless
  indoor channels: a communication-theoretic view,'' \emph{{IEEE} J. Sel. Areas
  Commun.}, vol.~20, no.~9, pp. 1613--1627, Dec. 2002.

\bibitem{JohShuPet:J07}
K.~Johan, W.~Shurjeel, A.~Peter, T.~Fredrik, and M.~A. F., ``A
  measurement-based statistical model for industrial ultra-wideband channels,''
  \emph{{IEEE} Trans. Wireless Commun.}, vol.~6, no.~8, pp. 3028--3037, Aug.
  2007.

\bibitem{DamDavWin:J08}
J.~D. B., D.~Davide, and W.~M. Z., ``Position error bound for uwb localization
  in dense cluttered environments,'' \emph{{IEEE} Trans. Aerosp. Electron.
  Syst.}, vol.~44, no.~2, pp. 613--628, Jul. 2008.

\bibitem{KleMih:J18}
B.~Klemen and M.~Mihael, ``Improving indoor localization using convolutional
  neural networks on computationally restricted devices,'' \emph{{IEEE}
  Access}, vol.~6, pp. 17\,429--17\,441, Mar. 2018.

\bibitem{HaoOzgMik:J19}
S.~Haoran, K.~A. Ozge, M.~Mike, V.~Harish, and H.~Mingyi, ``Deep learning based
  preamble detection and toa estimation,'' Dec. 2019, pp. 1--6.

\bibitem{AngMazSalFanChi:J20}
S.~Angarano, V.~Mazzia, F.~Salvetti, G.~Fantin, and M.~Chiaberge, ``Robust
  ultra-wideband range error mitigation with deep learning at the edge,''
  \emph{ArXiv}, vol. abs/2011.14684, May 2020.

\bibitem{LiMazShe:C22}
Y.~Li, S.~Mazuelas, and Y.~Shen, ``\textnormal{Deep Generative Model for
  Simultaneous Range Error Mitigation and Environment Identification},'' in
  \emph{Proc. IEEE Global Telecomm. Conf.}, 2022, \textnormal{To Appear}.

\bibitem{LiMazShe:C22_2}
------, ``\textnormal{A Deep Learning Approach for Generating Soft Range
  Information from RF Data.}'' in \emph{Proc. IEEE Global Telecomm. Conf.
  Workshop}, 2022, \textnormal{To Appear}.

\bibitem{Zhou:J18}
Z.-H. Zhou, ``A brief introduction to weakly supervised learning,''
  \emph{National science review}, vol.~5, no.~1, pp. 44--53, 2018.

\bibitem{YeHuaWan:J19}
X.~Ye, S.~Huang, Y.~Wang, W.~Chen, and D.~Li, ``Unsupervised localization by
  learning transition model,'' \emph{Proceedings of the ACM on Interactive,
  Mobile, Wearable and Ubiquitous Technologies}, vol.~3, no.~2, pp. 1--23,
  2019.

\bibitem{JunJiaJie:J12}
J.~J. Pan, S.~J. Pan, J.~Yin, L.~M. Ni, and Q.~Yang, ``Tracking mobile users in
  wireless networks via semi-supervised colocalization,'' \emph{{IEEE} Trans.
  Pattern Anal. Mach. Intell.}, vol.~34, no.~3, pp. 587--600, Aug. 2012.

\bibitem{MazPer:C19}
S.~Mazuelas and A.~P{\'e}rez, ``General supervision via probabilistic
  transformations,'' in \emph{24th European Conference on Artificial
  Intelligence-ECAI 2020}, Aug. 2020, pp. 1348--1354.

\bibitem{WymMarGifWin:J12}
H.~Wymeersch, S.~Maran{\`o}, W.~M. Gifford, and M.~Win, ``A machine learning
  approach to ranging error mitigation for {UWB} localization,'' \emph{{IEEE}
  Trans. Commun.}, vol.~60, pp. 1719--1728, Apr. 2012.

\bibitem{MazConAllWin:J18}
S.~Mazuelas, A.~Conti, J.~C. Allen, and M.~Z. Win, ``Soft range information for
  network localization,'' \emph{{IEEE} Trans. Signal Process.}, vol.~66,
  no.~12, pp. 3155--3168, Jun. 2018.

\end{thebibliography}

\end{document}